# Leveraging Imperfection with MEDLEY A Multi-Model Approach Harnessing Bias in Medical AI

Farhad Abtahi[1,2,3], Mehdi Astaraki[1], Fernando Seoane [1,3,4,5]

[1]Department of Clinical Science, Intervention and Technology, Karolinska Institutet, 17177 Stockholm, Sweden
[2]Department of Biomedical Engineering and Health System, School of Engineering Sciences in Chemistry, Biotechnology and Health, KTH Royal Institute of Technology, 14157 Huddinge, Sweden
[3]Department of Clinical Physiology, Karolinska University Hospital, 17176 Stockholm, Sweden
[4]Department of Textile Technology, Faculty of Textiles, Engineering and Business Swedish School of Textiles, University of Borås, 503 32 Borås, Sweden
[5]Department of Medical Technologies, Karolinska University Hospital, 141 57 Huddinge, Sweden

**Abstract**

Bias in medical artificial intelligence is conventionally viewed as a defect requiring elimination. However, human reasoning inherently incorporates biases shaped by education, culture, and experience, suggesting their presence may be inevitable and potentially valuable. We propose MEDLEY (Medical Ensemble Diagnostic system with Leveraged diversitY), a conceptual framework that orchestrates multiple AI models while preserving their diverse outputs rather than collapsing them into a consensus. Unlike traditional approaches that suppress disagreement, MEDLEY documents model-specific biases as potential strengths and treats hallucinations as provisional hypotheses for clinician verification. A proof-of-concept demonstrator was developed using over 30 large language models, creating a minimum viable product that preserved both consensus and minority views in synthetic cases, making diagnostic uncertainty and latent biases transparent for clinical oversight. While not yet a validated clinical tool, the demonstration illustrates how structured diversity can enhance medical reasoning under clinician supervision. By reframing AI imperfection as a resource, MEDLEY offers a paradigm shift that opens new regulatory, ethical, and innovation pathways for developing trustworthy medical AI systems.

## 1   Introduction

In the seven decades since Artificial Intelligence (AI) was first defined at Dartmouth, we have witnessed a rapid rise in attention over the past few years, mainly driven by public access to large language models (LLMs) and the push to integrate them into nearly every aspect of daily life. AI has rapidly expanded into medicine, with over a thousand FDA-cleared tools already used across various domains, including imaging, diagnostics, and workflow support. LLMs have accelerated expectations for clinical integration, offering capabilities such as summarizing patient records, generating reports, and assisting in diagnosis and treatment planning. However, despite their fame, no LLM-based system has been reviewed or approved by the FDA for clinical use  [1]. This gap reflects both the promise of these models and the unresolved risks they pose.

Bias, systematically favouring or disadvantaging certain groups, remains a pervasive challenge [2, 3]. Hallucinations, where models generate ungrounded outputs, are especially concerning in clinical contexts  [4]. Moreover, the "black-box" nature of deep learning systems complicates accountability and trust  [5]. These limitations are well documented in biomedical and computer science literature, highlighting the ethical stakes of deploying AI in high-risk domains such as health care [6, 7].

Regulators are responding with layered frameworks. The FDA emphasizes a lifecycle, risk-based approach to AI oversight, while the EU's Medical Device Regulation (MDR), In Vitro Diagnostic Regulation (IVDR), and AI Act designate medical AI systems, especially those used for diagnosis or treatment, as "high-risk," demanding bias mitigation, transparency, and human oversight. The EU has further advanced this by developing structured AI evidence pathways to operationalize trustworthy AI in healthcare, a roadmap for demonstrating safety, efficacy, and accountability across an AI system's lifecycle [8]. These guidelines aim to standardize documentation and validation evidence, ensure robust model performance across diverse populations, and support traceability, evaluation, and post-market monitoring.

These concerns echo a deeper question of responsibility: when an AI system fails, who is accountable? Clinicians often feel they remain ultimately liable, even when decisions are influenced by opaque AI systems [9]. Philosophical analyses describe this as a "responsibility gap," where neither clinicians, developers, nor institutions can be blamed for AI-driven harm [10]. Current consensus holds that AI models must function as decision-support systems, with the final decision remaining with the human expert in the loop. However, for this responsibility to be meaningful, clinicians must understand the basis of AI outputs. Explainable AI has been proposed as one solution. However, research shows that when explanations are overly persuasive, they can exacerbate automation bias, encouraging clinicians to over-trust outputs even when incorrect [11]. Similarly, recent analyses of large language models reveal that their behavior can echo human cognitive shortcuts, similar to Kahneman's "System 1" fast thinking, where seemingly intuitive but shallow reasoning leads to hallucinations or over-confident errors [12]. Together, these insights highlight the limits of relying on explanation alone and the need for frameworks that preserve diversity of perspectives and explicitly surface uncertainty.

This paper introduces MEDLEY (Medical Ensemble Diagnostic system with LEveraged diversitY) as an abstract framework for bias-aware and hallucination-tolerant medical AI. MEDLEY is a conceptual paradigm that treats bias and potentially hallucination as structured resources for decision support. Unlike traditional ensembles that collapse multiple outputs into a single answer, MEDLEY emphasizes diversity, transparency, plurality, and context by orchestrating various models in parallel. This conceptual framework applies to different domains; however, a proof-of-concept system was developed to demonstrate its feasibility. This demonstrator utilized over 30 LLMs across clinical cases to investigate diagnostic variability, bias attribution, and overall system performance. We then discuss how MEDLEY, as a conceptual approach, has broader implications for explainability, ethics, regulation, and innovation in medical AI.

## 2  Method

The MEDLEY framework is guided by four principles: *diversity*, *transparency*, *plurality*, and *context*. Unlike traditional ensemble methods that aggregate outputs into a single prediction, MEDLEY deliberately preserves multiple model outputs and their bias profiles to enrich clinical reasoning. *Diversity* ensures models with heterogeneous training protocols and/or learning algorithm architectures are included. *Transparency* requires documentation of provenance and limitations. *Plurality* emphasizes preserving distinct outputs rather than resolving conflicting predictions by highlighting the consensus. *Context* ensures clinicians can interpret results relative to patient-specific factors [13, 14]. Together, these principles reframe bias and hallucination not as defects but as resources for structured decision support.

Ensemble learning has historically been framed around error reduction. Classical methods, such as bagging and boosting, aggregate model outputs to suppress variance, while Bayesian ensembles integrate predictions into a single posterior-weighted distribution. Mixture-of-experts frameworks go further, using a gating network to route inputs to the "most appropriate" expert, collapsing plurality into an optimized pathway [15]. In all these paradigms, disagreement is treated as noise, and bias is managed implicitly through weighting or selection. While highly effective for benchmark accuracy, these approaches obscure the diversity of perspectives that may carry clinical significance. Table 1 contrasts classical ensemble paradigms with MEDLEY's approach.

*Table 1. Comparative paradigms in ensemble learning.*

| ASPECT | TRADITIONAL ENSEMBLE (BAGGING / BOOSTING) | BAYESIAN ENSEMBLE | MIXTURE-OF-EXPERTS | MEDLEY PARADIGM |
|---|---|---|---|---|
| AIM | Maximize predictive accuracy by collapsing outputs into a single result | Compute weighted posterior average across models | Route input to most relevant "expert" via gating network | Provide multiple diverse perspectives for clinician synthesis |
| OUTPUT | Single aggregated prediction (e.g., voting, weighted sum) | Probabilistic single prediction with posterior weighting | Single optimized output from selected expert(s) | Multiple outputs with provenance and bias annotations |
| TREATMENT OF DISAGREEMENT | Treated as noise/variance to be minimized | Down-weights outlier models | Suppressed by gating | Preserved as structured signal of diagnostic uncertainty |
| BIAS HANDLING | Attempt to cancel/eliminate bias statistically | Treated as prior or likelihood artifact | Hidden within gating decisions | Document, contextualize, and leverage bias transparently |
| HALLUCINATIONS / ERRORS | Suppressed as error | Integrated probabilistically | Filtered via gating | Used as speculative hypotheses with human verification |
| CLINICIAN COGNITION | Risk of automation bias; anchoring on single output | May obscure minority perspectives under averaged posterior | Reinforces dominant pathway chosen by gate | Encourages deliberation; surfaces disagreement patterns |
| ANALOGY | Averaged poll result/ Single symphony performance | Weighted belief distribution/ Single symphony performance | Referral to a single specialist/ Single symphony performance | Consultation panel / musical medley of distinct voices |

*Note: Classical ensembles (bagging/boosting), Bayesian ensembles, and mixture-of-experts aim to maximize predictive accuracy by collapsing multiple models into a single optimized output. In contrast, MEDLEY reframes ensemble learning for clinical contexts: disagreement is preserved as a marker of diagnostic uncertainty, and bias is treated as transparent specialization rather than error to be eliminated. This shift moves optimization away from accuracy alone toward structured diversity, provenance, and interpretability for human-in-the-loop decision making.*

MEDLEY diverges from these traditional approaches by preserving rather than collapsing disagreement and optimizing for diagnostic diversity rather than aggregate accuracy. This conceptual shift reframes ensemble learning for high-stakes clinical contexts, where visibility of uncertainty, population-specific knowledge, and plural perspectives may be more valuable than marginal gains in predictive accuracy.

**Bias Taxonomy**

Bias in medical AI arises from multiple sources, including historical, demographic, measurement, and deployment-related factors. Such biases are pervasive in systems trained on electronic health records (EHRs), [2], and are well-documented across the ML lifecycle [14]. Rather than treating bias solely as a liability, MEDLEY interprets it as a form of specialization that can provide clinically relevant insight. For example, geographic biases may improve recognition of regional diseases, while temporal biases capture evolving diagnostic practices. By explicitly documenting provenance and maintaining model plurality, MEDLEY allows clinicians to weigh outputs with awareness of their limitations. Table 2 summarizes nine categories of bias and the orchestration strategies MEDLEY applies, while Figure 1 visualizes their alignment with the four guiding principles [7].

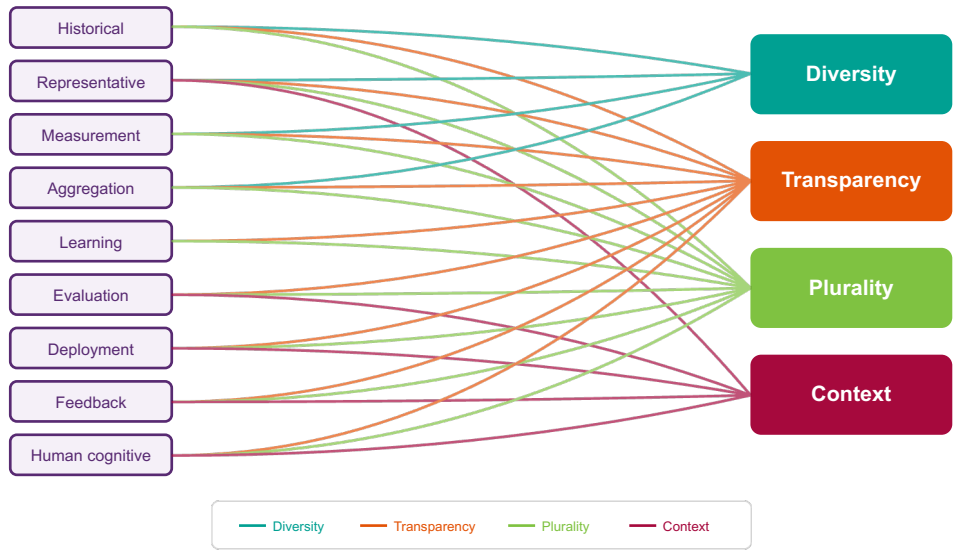

**Figure 1. Bias categories and their alignment with MEDLEY principles.** Nine types of bias common in medical AI are mapped against MEDLEY's four guiding principles. Rather than eliminating bias, MEDLEY treats bias as specialization when transparently documented, allowing context-sensitive use of diverse model perspectives.

The MEDLEY pipeline translates these principles into a three-stage orchestration architecture (Figure 2). In the first stage, multiple heterogeneous models run in parallel on the same patient input following the MEDLEY principles. The design ensures both intra-modality and cross-modality representation. For example, several convolutional neural networks may analyse a chest CT trained on datasets from different regions and/or institutions. At the same time, multiple language models with distinct training corpora, fine-tuning strategies, and cultural provenance could process EHR data. Similarly, competing statistical or machine learning models may interpret structured laboratory values optimized for different populations.

The second stage orchestrates these outputs through comparative analysis, synthesis, and bias attribution. Instead of collapsing predictions into a single consensus, as in traditional ensemble learning [15], MEDLEY evaluates patterns of agreement and disagreement, mapping them to the documented characteristics of each model. Disagreement is not discarded but treated as a marker of diagnostic uncertainty or population-specific specialization. A synthesis layer then organizes the outputs into a structured diagnostic report, explicitly distinguishing between consensus findings, plausible alternatives, and minority predictions that may highlight rare conditions.

The third stage presents these orchestrated outputs to clinicians in an interpretable format. Consensus predictions are highlighted but accompanied by divergent outputs, each annotated with provenance information such as training population, institutional source, or known performance limitations. For instance, if only a regional model identifies Familial Mediterranean Fever, this is surfaced with transparent notes about the model's geographic training data. The interface supports active clinical reasoning while explicitly showing uncertainty and bias profiles, while counteracting automation bias [5].

Together, this pipeline shifts the goal of ensemble learning from maximizing accuracy through uniformity to enhancing insight through structured diversity. By preserving both consensus and disagreement, MEDLEY enables clinicians to navigate multiple perspectives rather than over-trusting a single output, aligning with ethical and regulatory expectations for high-risk medical AI [7, 13].

*Table 2. Taxonomy of bias in clinical AI and MEDLEY's orchestration strategies.*

| BIAS CATEGORY | DEFINITION | CLINICAL EXAMPLE | MEDLEY RESPONSE |
|---|---|---|---|
| HISTORICAL | Outdated concepts embedded in data | Old diagnostic criteria miss recent discoveries | MEDLEY applies the principle of diversity by incorporating models trained on temporally distinct cohorts and the principle of transparency by leveraging documented temporal provenance, thereby capturing the evolution of medical knowledge. |
| REPRESENTATION | Training-population mismatch | Western-heavy datasets; race/gender imbalance | MEDLEY enacts diversity by selecting models trained on demographically and geographically varied populations and plurality by preserving subgroup-specific outputs that reflect meaningful variation in disease prevalence. |
| MEASUREMENT | Proxy variables oversimplify or misrepresent | ZIP code as genetic proxy; race coded as biology; inconsistent annotation practices | MEDLEY advances transparency by exposing proxy definitions and context by allowing clinicians to interpret model outputs in light of differing measurement conventions and their appropriateness for individual patients. |
| AGGREGATION | Loss of subgroup nuance | Pooling paediatrics/adult data or male/female physiology | MEDLEY promotes plurality by retaining subgroup-specific models and context by enabling clinicians to weight such models according to patient characteristics. |
| LEARNING | Algorithmic choices amplify disparities | Majority-class overweighting; underfitting rare subgroups | MEDLEY leverages diversity by integrating models with heterogeneous architectures and inductive biases, and plurality by preserving their distinct outputs rather than collapsing them. |
| EVALUATION | Test data unrepresentative | Benchmarks omit rare diseases; validation lacks subgroup diversity | MEDLEY applies diversity by incorporating models validated across heterogeneous benchmarks and transparency by reporting subgroup-specific performance and coverage gaps. |
| DEPLOYMENT | Use outside intended scope | Model validated in men applied in women's clinics | MEDLEY ensures transparency by surfacing models' documented scope and intended use and context by enabling clinicians to judge the appropriateness of applying outputs in a given setting. |
| HUMAN FACTORS | Clinician reliance or bias in interpreting AI | Automation bias; stereotype-driven interpretation | MEDLEY embodies plurality by presenting multiple model outputs and context by requiring clinicians to adjudicate among them, thereby counteracting over-reliance on a single system. |
| FEEDBACK | Model use influences future data, creating self-reinforcing bias | AI over-diagnoses a condition in one group, leading to inflated incidence in future training data | MEDLEY employs transparency by tracing the provenance of models and datasets across time, plurality by maintaining independent perspectives to avoid reinforcing single-model errors, and context by requiring clinicians to assess whether observed patterns reflect true prevalence or feedback loops. |

*Note: Bias categories include historical, representation, measurement, aggregation, learning, evaluation, deployment, human cognitive, and feedback loops. MEDLEY operationalizes four principles, diversity, transparency, plurality, and context, to convert these biases into sources of complementary insight.*

## 2.1 Clinical Implementation Scenario

To demonstrate the functionality of MEDLEY in an LLM-based context, we developed a synthetic, yet clinically plausible case study. The case involves a 45-year-old male of Middle Eastern origin presenting with a constellation of symptoms, including chest pain, fatigue, and intermittent fever. This illustrative example showcased the system's practical application and ability to process complex clinical information. Four language models representing different training populations produced divergent hypotheses: viral myocarditis, Familial Mediterranean Fever, anxiety disorder, and pericarditis. The synthesis engine identified an inflammatory cardiac process as a unifying theme while preserving the regional model's recognition of Familial Mediterranean Fever. This illustrates how MEDLEY surfaces population-specific conditions that could be overlooked in single-model approaches (Table 3).

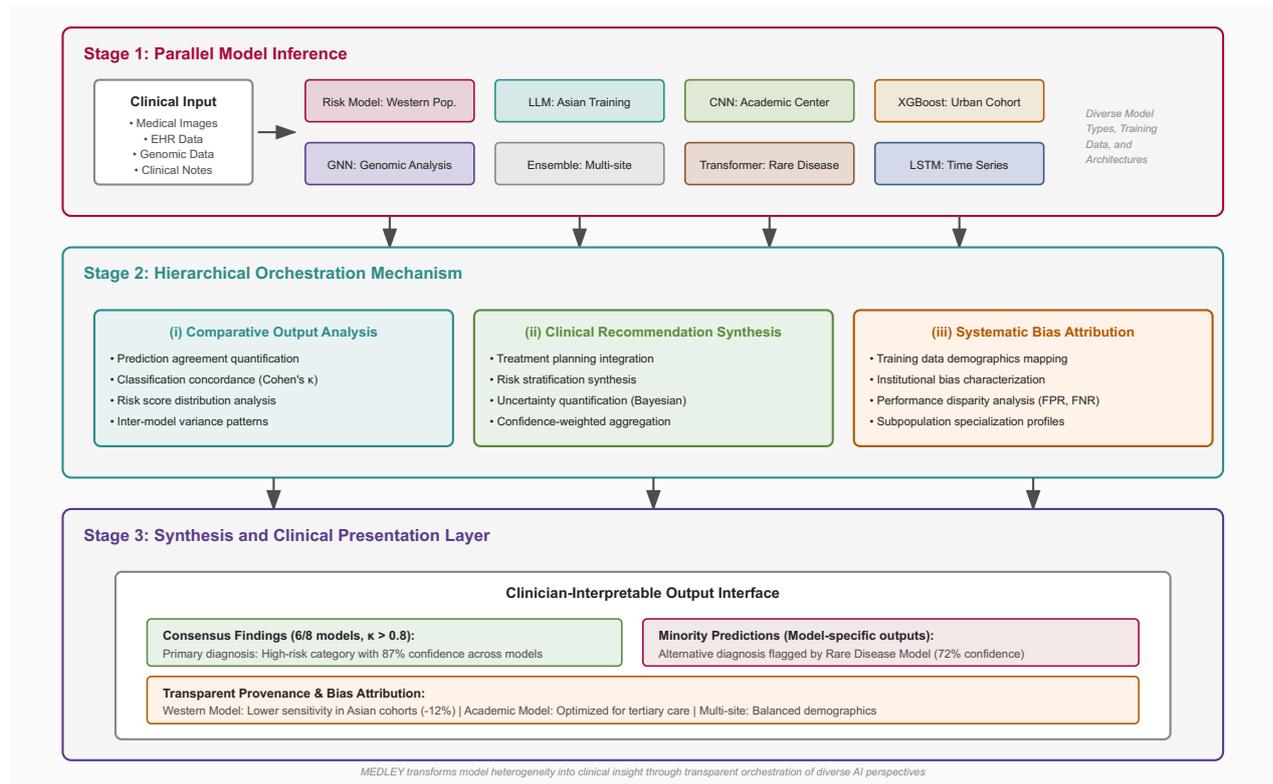

**Figure 2. The MEDLEY pipeline architecture.** Stage 1: diverse models within and across modalities (e.g., multiple CNNs or transformers for imaging, multiple LLMs for text/EHR, statistical or classification machine learning models for tabular data) are run in parallel. Stage 2: a hierarchical orchestration layer conducts comparative analysis, synthesizes recommendations, and attributes outputs to known bias profiles. Stage 3: results are presented to clinicians with both consensus and minority outputs preserved, annotated by provenance. This design turns model heterogeneity into structured clinical insight.

*Table 3. Clinical impact comparison of single model versus MEDLEY outputs for the hypothetical synthetic case*

| ASPECT | TRADITIONAL SINGLE-MODEL SYSTEM | MEDLEY MULTI-MODEL ENSEMBLE |
|---|---|---|
| PRIMARY OUTPUT | Single diagnosis: Viral myocarditis (78%) | Multiple differential diagnoses with explicit uncertainty |
| CLINICAL PRESENTATION | Standard algorithmic interpretation | Context-aware, multi-perspective analysis |
| REGIONAL CONSIDERATIONS | Not identified | Familial Mediterranean Fever flagged (75% confidence) |
| PHYSICIAN INTERFACE | Binary: Accept or Reject | Active selection from ranked differential |
| DIAGNOSTIC WORKUP | Standard cardiac protocol only | Comprehensive: cardiac + genetic (MEFV) + inflammatory markers |
| POTENTIAL OUTCOME | Missed genetic condition; delayed diagnosis | Early identification of population-specific disease |
| CLINICAL DECISION SUPPORT | Promotes automation bias | Promotes active clinical reasoning |
| HEALTH EQUITY IMPACT | May perpetuate diagnostic disparities | Addresses underrepresentation in training data |

### 2.2  Proof-of-concept LLM Demonstrator

To illustrate the feasibility of the conceptual MEDLEY framework, we implemented a minimum viable product (MVP) orchestrator integrating more than 30 LLMs from diverse geographic and architectural origins; see Figure 3. This prototype was designed to demonstrate orchestration mechanics, not to establish clinical validity or safety. All cases were synthetic, and the system should be understood as an illustrative demonstration rather than a deployable clinical tool. Each model was queried with identical text-based clinical cases and returned structured outputs including candidate diagnoses, International Classification of Diseases (ICD)-10 codes, and confidence scores. The orchestrator stratified results into three categories: Primary (≥30% of models), Alternative (10–29%), and Minority (<10%).

This proof-of-concept was deliberately limited to the text domain, demonstrating how MEDLEY can preserve both consensus and minority perspectives. Minority outputs often corresponded to rare or region-specific conditions, while consensus patterns reflected mainstream diagnoses. This architecture thus provides clinicians with a structured spectrum of diagnostic possibilities, rather than a collapsed single prediction.

## 3  Results

This proof-of-concept validates the computational tractability of the MEDLEY framework and demonstrates its potential as an extensible architecture adaptable to diverse medical AI modalities beyond language models. Supplement I presents an example generated report; additional reports are available on GitHub/website

### 3.1 Diagnostic Diversity

Across 12 representative synthetic diagnostic cases, model consensus rates varied widely, ranging from as low as 48% (Dementia with Lewy Bodies) to as high as 95% (Behçet's Disease, Sarcoidosis). Cases with lower consensus were enriched for rare or region-specific conditions, while higher consensus cases reflected well-established diagnoses.

Table 3 illustrates representative examples, showing how divergent outputs revealed diagnostic uncertainty. For instance, in IgA nephropathy, the ensemble produced 58 distinct alternative differentials, highlighting the ability of MEDLEY to preserve diagnostic breadth. In contrast, McArdle disease or Wilson's disease showed stronger consensus, with minority outputs contributing additional but less frequent hypotheses.

*Table 3. Representative diagnostic diversity in synthetic clinical cases.*

| CLINICAL CASE | PRIMARY DIAGNOSIS | ICD-10(-CM) CODES | MODEL CONSENSUS (%) | ALTERNATIVE DIAGNOSES (N) | NOTABLE INSIGHT |
|---|---|---|---|---|---|
| FAMILIAL MEDITERRANEAN FEVER | Familial Mediterranean Fever | M04.1, E85.0 | 63 | 29 | Regional models recognized FMF more reliably |
| DEMENTIA WITH LEWY BODIES | Dementia with Lewy Bodies | G31.83 | 48 | 3 | Some models generated speculative symptom links |
| METHAMPHETAMINE-INDUCED PSYCHOSIS | Methamphetamine-induced psychotic disorder | F15.959 | 75 | 3 | Moderate consensus achieved |
| MCARDLE DISEASE | McArdle Disease (Glycogen Storage Disease V) | E74.04 | 72 | 3 | Rare disease - variable model recognition |
| MANGANESE TOXICITY AND PARKINSON | Parkinson's Disease | G20, T56.0X1A | 65 | 4 | Environmental exposure case |
| SARCOIDOSIS | Congestive Heart Failure | I50.9, D86.0 | 95 | 3 | Strong consensus achieved |
| BEHÇET'S DISEASE | Deep Vein Thrombosis (DVT) | I82.90, M35.2 | 95 | 3 | Strong consensus achieved |
| ACUTE PORPHYRIA | Meningitis/Meningoencephalitis | G03.9, E80.20 | 65 | 4 | Moderate consensus |
| HEMOCHROMATOSIS | Atrial Fibrillation | I48.91, E83.110 | 85 | 3 | Strong consensus achieved |
| WILSON'S DISEASE | Post-Traumatic Stress Disorder | F43.10, E83.01 | 90 | 3 | Strong consensus achieved |
| AUTOIMMUNE ENCEPHALITIS | G6PD Deficiency with Hemolytic Anemia | D55.0, G04.81 | 85 | 3 | Strong consensus achieved |
| IGA NEPHROPATHY | IgA Nephropathy | N02.8, N03.9 | 53 | 58 | High diagnostic diversity; implausible proposals flagged |

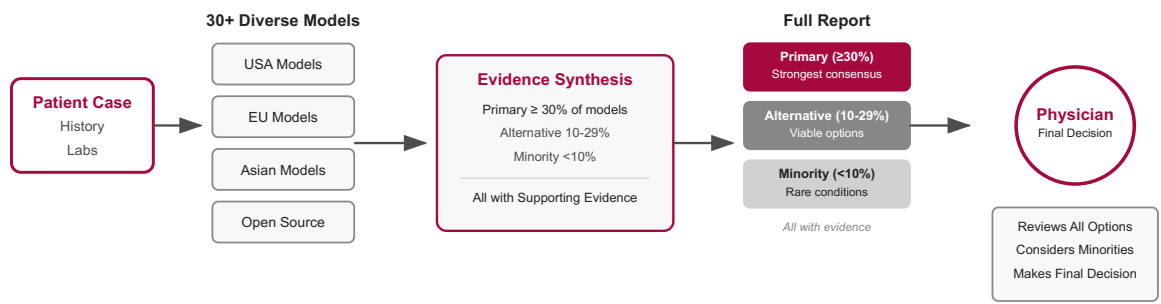

**Figure 3. Proof-of-concept demonstrator with over 30 LLMs.** Models were orchestrated in parallel and their outputs stratified into primary (≥30%), alternative (10–29%), and minority (<10%) diagnoses. Consensus diagnoses reflected mainstream conditions, while minority results highlighted rare or regional diseases. Rather than collapsing disagreement, MEDLEY preserved and structured it, demonstrating the feasibility of large-scale orchestration while enhancing transparency for clinicians.

Beyond case-specific variation, comprehensive analysis of model behaviour across all cases revealed six key performance dimensions (Table 4). Cost considerations showed that free models achieved comparable consensus alignment to paid alternatives (58.1% vs 57.8%), demonstrating that diagnostic quality is not determined by commercial accessibility. Geographic distribution analysis revealed US model dominance (65%), with limited representation from Europe (10%), China (16%), and other regions (9%), highlighting the need for broader international representation in clinical AI ensembles.

Confidence expression patterns showed that almost all models appropriately expressed uncertainty markers, indicating proper calibration to case complexity for safe clinical communication. Diagnostic breadth averaged 8.5 unique diagnoses per case (range: 1-59), with an inverse correlation to consensus; high consensus cases (>80%) generated fewer alternatives (average four diagnoses). In comparison, low consensus cases (<60%) produced richer differential lists (average 32 diagnoses).

Individual model performance identified top consensus leaders, including DeepSeek Chat (83.3%), Claude Opus (83.3%), and Mistral Large (66.7%), with both free and paid models achieving high consensus rates. Finally, consensus categorization distributed models across three participation levels, demonstrating that disagreement among models preserves diverse clinical perspectives rather than representing mere noise.

The ensemble analysis revealed distinct patterns of consensus participation that inform clinical interpretation strategies:

- High-consensus models (≥60% agreement, n=3): DeepSeek, Mistral Large, and Claude Opus aligned closely with mainstream clinical reasoning, providing reliable diagnostic anchors
- Moderate-consensus models (30-60% agreement, n=9): Including Qwen, GPT-4o, and Grok variants, these contributed balanced reasoning, generating both consensus and alternative diagnoses.
- Low-consensus models (<30% agreement, n=10): Including Gemini Flash and smaller parameter models, these acted as contrarian voices, more likely to propose rare or unconventional conditions.

These patterns demonstrate that disagreement among models might not be noise but an informative signal of diagnostic uncertainty, enriching the clinical reasoning process.

*Table 4. Comparative Analysis of models performance and characteristics across all cases.*

| MODEL CATEGORY | PERFORMANCE CHARACTERISTIC | KEY FINDING | CLINICAL VALUE |
|---|---|---|---|
| FREE VS PAID MODELS | Consensus alignment: Free 58.1% vs Paid 57.8% | No significant performance difference | Cost-effective implementation possible |
| GEOGRAPHIC DISTRIBUTION | US models dominate (65%), Europe (10%), China (16%), Other (9%) | Limited geographic diversity in current ensemble | Need for broader international representation |
| CONFIDENCE EXPRESSION | 95-100% of models express uncertainty markers | Appropriate calibration to case complexity | Safe uncertainty communication |
| DIAGNOSTIC BREADTH | Average 8.5 unique diagnoses per case (range: 1-59); inversely correlates with consensus | High consensus (>80%): avg 4 diagnoses; Low consensus (<60%): avg 32 diagnoses | Diagnostic uncertainty drives richer differentials |
| INDIVIDUAL MODEL LEADERS | Top consensus: DeepSeek Chat (83.3%), Claude Opus (83.3%), Mistral Large (66.7%) | Both free and paid models achieve high consensus | Quality not determined by cost |
| CONSENSUS CATEGORIES | High (≥60%): 3 models, Moderate (30-60%): 9 models, Low (<30%): 10 models | Wide variation in consensus participation | Diverse perspectives maintained |

Note: Confidence and uncertainty markers were analyzed to assess model calibration. Uncertainty: "possibly," "might be," "unclear," "cannot rule out." Confidence: "definitely," "certainly," "pathognomonic," "highly suggestive.

### 3.2 Bias Attribution and Disagreement Patterns

The ensemble revealed systematic biases that shaped diagnostic reasoning, which MEDLEY could surface and contextualize (Table 5). These bias patterns often contradicted intuitive expectations about model behavior and training origins. Geographic patterns showed unexpected results: European models demonstrated the lowest recognition rate for Mediterranean Fever (2.0 mentions/model) compared to US models (5.6 mentions/model), despite presumed regional familiarity. This suggests that geographic model origin alone does not predict better recognition of regional diseases, with training data curation appearing more influential than model provenance. Temporal biases favored historical over contemporary conditions, with AIDS/HIV appearing across multiple cases (24 mentions) despite only the homeless patient having relevant risk factors. At the same time, COVID-19 received zero mentions despite post-2020 model training. This indicates that training data recency might affect differential considerations in unexpected ways.

These findings arise from synthetic test cases and demonstrate MEDLEY's ability to make underlying biases visible. Confirmation of whether such patterns hold in clinical practice will require empirical validation with real-world datasets and patient populations. In this sense, the results highlight the framework's hypothesis-generating potential rather than serving as clinical evidence.

Confidence calibration proved appropriately responsive to case complexity, with uncertainty markers scaling from 5 in straightforward methamphetamine psychosis to 46 in complex IgA nephropathy. Diagnostic diversity showed an inverse correlation with consensus; high diversity cases, like IgA nephropathy (53% consensus, 58 alternatives), contrasted sharply with high consensus cases like Sarcoidosis and Behçet's disease (95% consensus, three alternatives each). Demographic anchoring was pronounced across relevant cases: substance use was mentioned 2.5 times per model for the homeless patient. In comparison, age factors appeared 1.6 times per model for the older woman with confusion. Treatment approaches showed no systematic bias toward over- or under-treatment, with models splitting relatively evenly between aggressive and conservative recommendations across different patient demographics.

These illustrative findings suggest that MEDLEY can surface and contextualize minority results and biases that would remain hidden, helping clinicians interpret diagnostic variation within a transparent framework.

*Table 5. Examples of Bias attribution in MEDLEY orchestration.*

| BIAS TYPE | MANIFESTATION | EXAMPLE FROM CASES | CLINICAL IMPLICATION |
|---|---|---|---|
| GEOGRAPHIC | Disease recognition shows limited correlation with model origin | In FMF case European models had lowest recognition (2.0 mentions/model) despite Mediterranean relevance; In Meningitis case: Chinese models showed preference for Asian diseases (Japanese encephalitis) | Geographic model origin alone does not predict better recognition of regional diseases; other factors like training data curation appear more influential |
| TEMPORAL | Historical conditions over-represented | AIDS/HIV mentioned in majority of cases (24 mentions) despite no relevant risk factors; Wegovy appears in Case 11 showing recent drug awareness | Training data recency affects differential considerations and treatment options |
| CONFIDENCE EXPRESSION | Model certainty varies with case complexity | Complex IgA nephropathy: 46 uncertainty vs 19 confidence markers; Straightforward methamphetamine psychosis: 41 uncertainty vs 5 confidence markers; Uncertainty appropriately increases with diagnostic difficulty | Models express calibrated confidence based on case ambiguity |
| DIAGNOSTIC DIVERSITY | Consensus inversely correlates with alternatives | High diversity: IgA nephropathy (53% consensus, 58 alternatives); High consensus (>90%): Sarcoidosis, Behçet's, Wilson's disease (95%, 95%, 91% consensus, 3 alternatives each); | Higher diagnostic uncertainty produces richer differential lists |
| DEMOGRAPHIC FOCUS | Social factors amplified in relevant cases | Homeless patient: Substance use mentioned 2.5x per model; Elderly woman with confusion: Age mentioned 1.6x per model; Teen swimmer with exercise intolerance: Lifestyle factors 1.4x per model | Demographic salience creates diagnostic anchoring patterns |
| TREATMENT APPROACH | Aggressive vs conservative split | Homeless with psychosis: 27 aggressive vs 26 conservative models; Tech executive from Bangladesh: 22 aggressive vs 21 conservative; Testing intensity varies: Bangladesh case had 21 models suggest extensive testing vs IgA nephropathy only 3 models | No systematic bias toward over- or under-treatment across models |

### 3.3 System Performance (Demonstrator)

The orchestrator successfully demonstrated the computational feasibility of coordinating more than 30 LLMs, with routine ensemble runs completing within minutes. Parallel querying and staged synthesis supported efficient throughput, and the failover system reliably redirected requests when token overflows or timeouts occurred. Claude 3.5 Sonnet was the primary synthesis engine by default, with GPT-4o and Gemini 2.0 Pro as backup options.

While technically functional, the prototype was constrained by API usage caps, occasional model unavailability, and variable latency, all of which introduced scalability limits. Despite these constraints, the demonstrator proved robust enough to manage large-scale orchestration. Its success validates the feasibility of the MEDLEY paradigm at a proof-of-concept level, though production-level deployment will require dedicated resources, standardized orchestration interfaces, and further empirical testing.

It is worth noting that employing over 30 LLMs for demonstrative purposes can be computationally intensive. In contrast, other practical applications can be processed on local infrastructures, such as using several computer vision models for image analysis or classical machine learning and statistical models for tabular data. This reduces the dependency on computationally expensive cloud systems, significantly lowering computational demands.

## 4 Discussion

In this study, we introduced MEDLEY, a conceptual framework and demonstrator that reimagines ensemble learning in medical AI. Unlike traditional ensemble methods, prioritizing predictive accuracy by collapsing outputs into a single consensus, MEDLEY preserves diagnostic diversity and makes disagreement visible through structured orchestration. By linking divergent outputs to known bias profiles, MEDLEY reconceptualizes model imperfection not as a limitation but as a valuable resource for enhanced clinical reasoning. Our demonstrative example, conducted within a synthetic yet clinically plausible scenario using large language models (LLMs), shows the technical feasibility of orchestrating more than 30 LLMs. The results indicate that this orchestration reveals systematic patterns of consensus, disagreement, and bias, insights that are typically obscured in single model systems. In this study, the findings should be viewed as illustrative of feasibility rather than evidence of clinical performance, highlighting the hypothesis-generating role of MEDLEY.

This study's demonstration of MEDLEY using LLMs serves as a proof-of-concept; however, the framework's principles extend far beyond this specific application. MEDLEY can be applied to diverse domains within medical AI, including classical machine learning for EHRs and tabular data, deep learning models for medical imaging, and biosignal analysis. Consider the field of medical image segmentation. Recent innovations in deep learning have led to the development of numerous models for applications like tumour segmentation [16]. While some studies suggest that ensemble techniques like STAPLE or majority voting can slightly improve performance [17], other research indicates that a single, robust model may still outperform these aggregated results [18]. This discrepancy underscores a fundamental limitation of traditional, accuracy-centric ensemble approaches. These methods prioritize maximizing overall performance and often obscure crucial disagreements among models.

For instance, in the MICCAI Brain Tumour Segmentation (BraTS) challenge, models are benchmarked on their ability to segment distinct tumour subregions [19]. While subregional segmentation is vital for biomarker studies, clinical radiation therapy (RT) often requires only the union of these regions for treatment planning purposes. Traditional ensemble methods that aggregate consensus on these subregions may overlook critical discrepancies, such as over- or under-segmentation, that could have clinical consequences. In RT, such discrepancies can be vital for avoiding radiation exposure to sensitive structures. In this regard, MEDLEY offers a conceptual alternative: making model disagreements visible and linking them to specific clinical considerations. This enables a nuanced investigation into the advantages and drawbacks of different segmentation methods. This contrasts with conventional ensembling, which aims solely to enhance aggregate performance without accounting for clinical relevance. Consequently, MEDLEY can help reveal which models are most appropriate for specific clinical applications, such as RT planning, by reframing imperfection as a resource for clinical reasoning.

### 4.1 Transforming AI Imperfection from Bug to Feature

The MEDLEY paradigm reorients rather than rejects dominant priorities in medical AI development. Efforts to reduce bias, mitigate error, and prevent hallucination remain ethically and clinically essential.

However, it is equally important to acknowledge that such imperfections can never be eliminated entirely. Cognitive shortcuts such as anchoring, availability, and overconfidence persist in human medicine even among experienced clinicians.

In a systematic review of 20 studies involving more than 6,800 physicians, Saposnik et al. found that these biases were associated with diagnostic inaccuracies in 36.5–77% of cases and with management errors in 71% of applicable studies [20]. These findings underscore that bias and error are intrinsic to clinical reasoning, not simply the result of inadequate training or oversight.

Acknowledging and managing cognitive fallibility within medicine are therefore central aspects of clinical practice, guiding the development of systematic frameworks to enhance decision-making. Clinical guidelines, first formalized in the 1990s by Eddy, represented a structured attempt to reduce variability in reasoning by making processes transparent and evidence-based [21]. Over time, these guidelines have been extended to incorporate individualized recommendations that adjust for patient risk, reflecting a continuous effort to manage, rather than eliminate, the complexities inherent in clinical decision-making [22].

Beyond guidelines, educational strategies have been developed to counter cognitive bias. Reflective prompting and metacognitive training are among the most studied, emphasizing the importance of acknowledging biases as a prerequisite for mitigation, even if their eradication remains elusive [21]. A scoping review by Monash Health found empirical support for reflective reasoning as an effective intervention for reducing diagnostic bias[22]. At the same time, Saposnik et al. have argued for integrating bias-awareness training directly into medical education to strengthen competency and safety [20]. Behavioral science adds further strategies: More wedge et al. propose targeted training, nudges, and incentive design as structured interventions to improve judgment, suggesting that even modest interventions can promote more deliberate reasoning [21]. These approaches demonstrate that medicine has sought to manage imperfection by turning insights from cognitive psychology into safeguards against error.

Perhaps the most enduring and pragmatic response has been the institutionalization of collective judgment. Tumor boards, case conferences, and second-opinion systems illustrate how medicine transforms heterogeneity into resilience. Multidisciplinary tumor boards, particularly in oncology, consistently improve staging accuracy, treatment planning, and patient outcomes [23]. Soukup et al. describe such forums as mechanisms that harness cognitive diversity and safeguard against individual bias through structured group deliberation [24].. MEDLEY draws inspiration from these institutional models, treating imperfection not as a flaw to be eliminated but as a resource to be orchestrated.

Biases may become constructive when explicitly documented and ethically governed, such as subgroup-specific models that improve recognition of endemic conditions. Likewise, hallucinations, if flagged as speculative, may broaden the diagnostic search space by prompting hypotheses that clinicians can evaluate and test. By preserving plurality, the MEDLEY paradigm may help counteract automation bias and anchoring, which are well-documented risks in human–AI collaboration[11, 25]. As shown in Table 3, these findings highlight that biases, though often framed solely as limitations, can be strategically leveraged in ensemble systems. The results serve to generate hypotheses and design principles for future empirical work.

Microsoft's MAI Diagnostic Orchestrator (MAI-DxO) [26] employs a model-agnostic orchestration framework that simulates a panel of physician-agents reasoning sequentially. Paired with OpenAI's o3 and other frontier models, it achieved up to 85.5% diagnostic accuracy in NEJM case simulations, over four times higher than generalist physicians (~20%), and reduced diagnostic costs by 20–70% compared with both physicians and baseline models. However, all outputs are ultimately generated by the same underlying LLM (e.g., GPT, Gemini, Claude). Diversity is simulated through role specialization, prompting strategies, or retrieval-augmented knowledge bases. Still, it is collapsed into a single optimized diagnostic pathway, with plurality as an intermediate step rather than a final deliverable.

Google's AMIE (Articulate Medical Intelligence Explorer) [27] exemplifies frontier conversational diagnostic reasoning performance. Trained via self-play in simulated physician–patient dialogues with automated feedback, AMIE outperformed primary care physicians in blinded OSCE-style evaluations across diagnostic accuracy, communication quality, management reasoning, and empathy. Yet, AMIE relies on a single optimized LLM (e.g., Gemini), with diversity introduced only via synthetic dialogue generation or role simulation during training. Its paradigm emphasizes natural conversational interaction rather than orchestrated plurality at inference time.

By contrast, MEDLEY orchestrates multiple heterogeneous models in parallel rather than simulating diversity within a single model. Each output is preserved with explicit bias profiles, embracing disagreement as clinically informative. This design emphasizes genuine intra- and inter-modality diversity through distinct systems working side by side. Table 6 provides a structured comparison, highlighting how MEDLEY's preserved plurality differs from the simulated or dialogic diversity of MAI-DxO and AMIE at the paradigm level.

*Table 6: Comparison of medical AI paradigms..*

| ASPECT | MEDLEY | MAI-DXO (MICROSOFT) | AMIE (GOOGLE) |
| --- | --- | --- | --- |
| CORE PHILOSOPHY | Embrace imperfection through diversity | Sequential physician-agent orchestration | Diagnostic dialogue expertise |
| MODEL STRATEGY | Multiple heterogeneous models in parallel, preserving intra- and inter-modality diversity | Multiple roles but same underlying LLM (e.g., GPT) | Single optimized LLM (e.g., Gemini) with self-play |
| OUTPUT | Multiple perspectives with explicit bias profiles | Single aggregated diagnostic pathway | Conversational diagnostic output |
| OPTIMIZATION TARGET | Diagnostic diversity & transparency | Accuracy & cost-effectiveness | Diagnostic accuracy, empathy, communication |
| CLINICAL ANALOGY | Consultation panel | Sequential specialist reasoning | Expert physician consultation |
| BIAS HANDLING | Explicit documentation of intra-modality variation and cross-modality specialization | Simulated diversity, not explicitly documented | Not explicitly addressed |
| HUMAN ROLE | Evaluate spectrum of opinions | Accept/reject aggregated recommendation | Engage in dialogue with AI |

*Note: Unlike MAI-DxO and AMIE, which simulate diversity through role-playing or dialogue within a single underlying LLM, the MEDLEY paradigm orchestrates multiple heterogeneous models both within and across modalities. This preserves divergent outputs and explicit bias profiles as final deliverables, aligning with tumor board–style consultation. Self-play is a training strategy in which a model simulates both sides of an interaction (e.g., patient and physician), generating synthetic dialogues evaluated with automated feedback. This method allows large-scale practice of diagnostic reasoning and communication skills without extensive human annotation, a technique initially developed in reinforcement learning for games such as chess and Go [28].*

## 4.2 The Risk of Digital Sophistry of Explainable AI in Medicine

Contemporary XAI enthusiasm assumes that the decision is trustworthy if a system articulates its reasoning. This assumption collapses with LLMs, which excel at generating polished narratives that sound clinically plausible while bearing little relation to actual computational processes [29, 30]. LLMs may provide diagnostic recommendations with impeccable-sounding rationales that are rhetorical afterimages rather than faithful reflections of internal reasoning [31], digital sophistry where persuasive speech masquerades as transparency [32].

MEDLEY sidesteps this paradox by refusing to rely on any single model's rhetorical skill, seeking reliability through a structured interplay of convergent and divergent perspectives across multiple models, a dialectic of outputs rather than reasoning performances.

## 4.3 Technical and Implementation Challenges

Realizing MEDLEY requires addressing technical, organizational, economic, regulatory, and ethical challenges. Parallel model execution introduces computational costs and latency, which can be mitigated through selective activation, complete ensembles for complex cases, and single models for routine tasks. Heterogeneous vendor ecosystems demand standardized interfaces to preserve diverse model contributions. Beyond technical constraints, adoption depends on organizational readiness, sustainable economics, and clear regulatory pathways. Table 7 summarizes implementation challenges and mitigation strategies.

## 4.4 Fostering Innovation Through AI Ecosystem Diversity

The MEDLEY paradigm challenges the monopolistic drift of the current AI marketplace, where pursuing a single "super model" concentrates power in a few corporations. MEDLEY creates viable niches for smaller, specialized vendors by valuing diversity over singularity. Startups focused on rare diseases, regional companies with local population data, or academic groups targeting underrepresented populations gain relevance within the ensemble, contributions devalued in single-model paradigms but essential in MEDLEY.

This ecosystem could conceptually mirror the app store model, enabling contributions from diverse developers. In a MEDLEY-style ecosystem, specialized dermatology models, regional cardiology predictors, or homegrown health models might coexist within and across domains, democratizing participation and potentially accelerating innovation. The vision represents a conceptual projection of ecosystem benefits rather than empirically validated outcomes, highlighting innovation pathways to be examined in future research.

*Table 7. Implementation challenges and potential mitigation strategies for the MEDLEY paradigm.*

| CHALLENGE CATEGORY | SPECIFIC ISSUES | PROPOSED SOLUTIONS |
|---|---|---|
| TECHNICAL | High computational cost and latency from intra- and inter-modality ensembles; interoperability across heterogeneous vendor systems | Selective activation of ensembles for complex cases; asynchronous result presentation; standardized orchestration interfaces |
| ORGANIZATIONAL | Workflow disruption; training requirements; cultural resistance among clinicians | Phased pilot implementations; simulation-based training; adoption led by local champions |
| ECONOMIC | Multiple vendor licenses; infrastructure costs; uncertain return on investment | Tiered service models (routine vs. complex cases); shared infrastructure; outcome-based pricing agreements |
| REGULATORY | Lack of ensemble-level validation frameworks; unclear allocation of liability; divergent international standards | Pilot programs in collaboration with regulators; transparent documentation and audit trails; efforts toward global harmonization |
| ETHICAL/SOCIAL | Patient trust in multi-model outputs; interpretability of intra-modality disagreements; ensuring equity across populations | Explicit disclosure in consent processes; public reporting of bias audits; inclusion of underrepresented populations in model development |

*Note: Running multiple models in parallel introduces costs, latency, interoperability, regulatory, and ethical hurdles. Crucially, MEDLEY requires orchestration of several models within each domain (e.g., multiple LLMs for text, multiple CNNs for imaging), which amplifies both the potential benefits and the challenges.*

## 4.5 Ethical Implications and Safeguards

MEDLEY reframes imperfection as a managed resource rather than aspiring to error elimination. Table 8 contrasts how this shifts the operationalization of biomedical ethics principles. Implementation requires informing patients that multiple systems contribute perspectives (autonomy), using redundancy for cross-validation (beneficence/non-maleficence), valuing population-specific models (justice), and surfacing all perspectives with bias attribution (transparency).

*Table 8. Comparison of ethical principles in traditional single-model AI versus the MEDLEY paradigm.*

| ETHICAL PRINCIPLE | TRADITIONAL SINGLE-MODEL APPROACH | MEDLEY PARADIGM | PRACTICAL IMPLEMENTATION |
|---|---|---|---|
| AUTONOMY | Generic patient consent to "AI use" | Patients informed that multiple models within and across domains contribute distinct perspectives | Standardized disclosure protocols |
| BENEFICENCE | Assumes a single best answer | Recognizes the value of divergent perspectives and diagnostic plurality | Present spectrum of diagnostic options |
| NON-MALEFICENCE | Strives to eliminate all error | Manages error through redundancy and cross-validation across models | Parallel review reduces risk of unchecked harm |
| JUSTICE | One-size-fits-all fairness metrics | Values subgroup-specific models; ensures minority perspectives are not averaged away | Incorporate underrepresented data sources |
| TRANSPARENCY | Explains a single output | Surfaces all perspectives, including intra-modality disagreements (e.g., between LLMs) and cross-modality variation (e.g., imaging vs. text) | Full reporting of model origins and rationale diversity |

*Note: MEDLEY treats imperfection and plurality as structured resources by preserving intra-modality disagreement and cross-modality perspectives. This reinterprets autonomy, beneficence, non-maleficence, justice, and transparency as collective safeguards rather than single-model obligations.*

### 4.6 Regulatory Pathways for MEDLEY

Adopting bias- and hallucination-aware strategies introduces profound regulatory challenges. If bias is intentionally embedded for constructive purposes (e.g., subgroup-focused models), regulators must ensure such practices promote equity without reinforcing stigma or exclusion. Policy frameworks increasingly recommend bias audits and fairness evaluations, such as calibration by subgroup and equality of error rates, as prerequisites for deployment. Regulators may also require explicit documentation of how sensitive attributes are used, preventing "bias-as-specialization" from becoming a loophole for discrimination.

Emerging prototypes illustrate feasible paths toward regulated deployment of hallucination-tolerant systems. In a team-based simulation of upper gastrointestinal bleeding management, [33] developed a CDSS that constrained LLM responses to structured risk models and guideline corpora, improving usability and trust without implementing automated deferral based on uncertainty thresholds. In radiology, CoRaX [34] functions as a post-interpretation "virtual second reader," integrating radiology reports, eye-gaze data, and imaging to detect and localize overlooked abnormalities. CoRaX corrected approximately 21% of simulated perceptual errors and generated referrals judged proper in 85% of interactions, underscoring the potential of collaborative AI to support rather than replace physician judgment.

These examples highlight that not every model in an ensemble requires independent clinical certification. Some models may be designed solely for orchestrated use under human oversight, paralleling existing device regulations where intended use varies by user expertise. MEDLEY requires regulatory pathways recognizing system-level ensemble certification, combined-use approval, and multi-level documentation, supported by audit logging and explicit oversight mechanisms.

### 4.7 Global Health and Equity Considerations

MEDLEY paradigm addresses the documented risk of "unbiased" universal models defaulting to majority populations and worsening care disparities. Evidence shows AI models trained on predominantly majority datasets produce substantial errors for minority patients [35], with performance disparities persisting even after bias assessments [36]. Conceptually, by orchestrating models trained on diverse populations within and across modalities, MEDLEY could preserve rather than average away minority perspectives. This feature may be crucial given that inadequate representation often leads to failures in predicting and treating conditions among underrepresented groups [37].

The framework transparency counters the dangerous assumption that AI systems are inherently objective [38]. By documenting each model's training population and surfacing disagreements, MEDLEY makes visible the biases that universal models obscure. Realizing this potential requires sustained investment in local model development, targeted data collection in underserved communities, data sovereignty protections, and regulatory standards that evaluate performance across demographic subgroups rather than aggregate metrics alone.

### 4.8 The Future of Human–AI Collaboration in Medicine

MEDLEY reframes the human-AI relationship toward augmentation rather than replacement, embodying the vision of man-computer symbiosis articulated by Licklider [39] and the framework for augmenting human intellect proposed by Engelbart [40]. Within this paradigm, clinicians function as orchestrators of diverse analytical perspectives, interpreting disagreements between models, e.g., radiology CNNs, reconciling conflicting risk scores from statistical models, and evaluating differential diagnoses generated by language models.

This approach, however, encounters fundamental cognitive constraints. According to Cognitive Load Theory [41], simultaneous presentation of multiple model outputs may exceed working memory capacity, potentially impairing rather than enhancing clinical decision-making. This phenomenon parallels the automation bias documented by Parasuraman and Riley [42], wherein excessive automated information paradoxically degrades human performance.

Successful implementation necessitates interfaces that respect cognitive processing limits [43] while preserving analytical diversity. Design principles must align with established clinical decision support guidelines, particularly the requirement to "anticipate needs and deliver in real time" while avoiding information overload [44]. Future research must determine optimal thresholds for ensemble activation to distinguish when multiple perspectives enhance versus impair clinical reasoning. The objective parallels the concept of Advanced Chess described by Kasparov [45], wherein human-machine teams demonstrate superior performance compared to either component independently. Without systematic consideration of cognitive load, MEDLEY risks violating core principles of adequate clinical decision support, specifically, the mandate to enhance rather than disrupt workflow [44]. The framework must therefore balance analytical comprehensiveness with cognitive feasibility to ensure that multiple perspectives enrich rather than burden clinical reasoning.

### 4.9 Future Research Directions

Building on this proof-of-concept, critical research pathways include: (1) empirical validation comparing single-model versus multi-model outcomes in clinical practice; (2) technical methods for optimal ensemble composition and context-sensitive bias weighting; (3) workflow strategies that minimize clinician cognitive load; (4) regulatory frameworks for ensemble-level approval and liability allocation; and (5) extension to multimodal data (imaging, laboratory values, physiological signals) while maintaining intra-modality diversity.

### 4.10 Out of Scope and Limitations

This work introduces a conceptual paradigm rather than validates a clinical tool. The demonstrator used only synthetic text-based cases as a minimal viable prototype to illustrate orchestration feasibility. No patient data, clinical accuracy, safety, or workflow utility were assessed. Technical implementation details and prompts are available in the repository (see Data and Code Availability).

The set of large language models in the ensemble lacked detailed bias documentation or training profiles. We inferred characteristics based on limited available information, e.g., release dates, parameter sizes, and country of origin, but these assumptions may not reflect actual training data composition or inherent biases. The models were general-purpose without integration of RAG-based systems that could access local clinical guidelines or institutional protocols essential for real-world implementation. Reliance on commercial APIs further limits reproducibility.

The results demonstrate computational tractability and orchestration mechanics, not real-world diagnostic performance. Future research must extend MEDLEY to multimodal data, integrate RAG-based systems, document actual model biases through empirical testing, evaluate clinical impact, and assess regulatory compliance before translation to practice.

## 5   Conclusion

This paper introduced MEDLEY, a conceptual framework for orchestrating imperfection in medical AI. MEDLEY preserves diagnostic plurality, unlike traditional ensembles that collapse outputs into a single answer. It makes bias visible, reframing imperfection as a resource rather than a liability, analogous to multidisciplinary tumour boards in clinical practice.

Our proof-of-concept demonstrated technical feasibility by orchestrating over 30 LLMs across synthetic diagnostic scenarios, revealing systematic patterns of consensus and disagreement. These findings illustrate orchestration mechanics and hypothesis-generating potential but do not validate clinical performance. Critical limitations include unknown model biases, lack of patient data, and absence of multimodal integration.

The paradigm suggests potential cost-effective pathways through existing specialized models, with a conceptual opportunity to foster ecosystem diversity for smaller contributors and to promote equity by valuing population-specific perspectives. However, translating this conceptual contribution to practice requires empirical validation, infrastructure investment, and careful governance to ensure safe implementation.

## Acknowledgments

This work was supported by EU grant 240037 from EIT Health and SMAILE (Stockholm Medical Artificial Intelligence and Learning Environments) core facility at Karolinska Institutet.

## Author contributions

F.A. conceived the MEDLEY paradigm, designed the framework, implemented the demonstrator, prepared the first draft of the manuscript, and developed the public GitHub repository. M.A. contributed to the evaluation and assessment of the paradigm and demonstrator. F.S. guided analysis, interpretation, and evaluation of the paradigm. All authors contributed to manuscript preparation, revision, and approval of the final version.

## Competing Interests

The authors declare no conflicts of interest. The research was conducted independently of any commercial relationships that could influence the work.

## Data and Code Availability

The complete MEDLEY implementation is openly available at https://github.com/ki-smile/medley under the MIT license. A live demonstrator will be hosted at https://www.medleyai.org, which provides open access to all evaluation cases without requiring API credentials. Users wishing to analyse new cases can do so through the Custom Analysis Interface by supplying an OpenRouter API key. The repository includes the orchestration engine, ensemble outputs for all evaluation cases, bias profiles, documentation, and Docker deployment scripts. All clinical cases are synthetic, and no patient data were used.

## AI Tools Disclosure

The authors used AI-based tools (e.g., large language models) for English proofreading and improving the readability of the manuscript. These tools were applied only to enhance clarity of expression and did not contribute to the conceptual content, data analysis, or scientific conclusions. The authors take full responsibility for the content of the manuscript.

## Ethics Statement

This research was conducted by the Declaration of Helsinki using synthetic cases.